\documentclass[10pt,twocolumn,letterpaper]{article}

\usepackage{iccv}
\usepackage{times}
\usepackage{epsfig}
\usepackage{graphicx}
\usepackage{amsmath}
\usepackage{amssymb}

\usepackage{adjustbox}
\usepackage{soul}
\usepackage{url}
\usepackage[utf8]{inputenc}
\usepackage[small]{caption}
\usepackage{booktabs}
\usepackage{algorithm}
\usepackage{algorithmic}
\usepackage{tikz}
\usepackage{calc}
\usepackage{pifont}
\usepackage{multirow}
\usepackage{multicol}
\usepackage{xcolor}
\usepackage{bbding}
\usepackage{wasysym}
\usepackage[labelsep=period]{caption}
\usepackage{subfig}

\def\HS{\hspace{\fontdimen2\font}}
\def\HSE{\HS\HS\HS\HS\HS\HS\HS\HS\HS\HS}

\usepackage[pagebackref=true,breaklinks=true,letterpaper=true,colorlinks,bookmarks=false]{hyperref}

\iccvfinalcopy 

\def\iccvPaperID{3388} 

\ificcvfinal\fi

\definecolor{darkblue}{HTML}{1F4E79}
\definecolor{lightblue}{HTML}{00B0F0}
\definecolor{salmon}{HTML}{FF9C6B}
\definecolor{golden}{HTML}{EDB304}
\definecolor{silvergray}{HTML}{BCBCBC}
\definecolor{pinky}{HTML}{FF3399}
\usetikzlibrary{shapes.geometric, arrows, backgrounds, fit, shapes.arrows,decorations.markings}
\tikzstyle{layer} = [rectangle, minimum width=1cm, minimum height=3cm, text centered, draw=black, fill=lightblue]
\tikzstyle{tinylayer} = [rectangle,minimum width=2cm, minimum height=15pt, text centered, draw=black, fill=silvergray]
\tikzstyle{tinythinlayer} = [rectangle,minimum width=2cm, minimum height=1pt, text centered, draw=black, fill=lightblue]
\tikzstyle{matrix} = [rectangle, minimum width=1.5cm, minimum height=1cm, text centered, draw=black, fill=golden]
\tikzstyle{arrow} = [->,>=stealth]
\tikzstyle{leftarrow} = [<-,>=stealth]
\tikzstyle{bigarrow}=[single arrow,draw=black,fill=black!10,minimum height=2cm]

\tikzstyle{vecArrow} = [-{Triangle[angle=40:2mm]}, ultra thick, black]

\begin{document}
\title{A Matrix-in-matrix Neural Network for Image Super Resolution}
\author{Hailong Ma \quad Xiangxiang Chu \quad Bo Zhang \quad Shaohua Wan \quad Bo Zhang\\
	Xiaomi AI\\
	Beijing, China \\
	{\tt\small \{mahailong, chuxiangxiang, zhangbo11, wanshaohua, zhangbo\}@xiaomi.com}
}

\twocolumn[
\begin{@twocolumnfalse}
\maketitle
\end{@twocolumnfalse}
]
\begin{abstract}
In recent years, deep learning methods have achieved impressive results with higher peak signal-to-noise ratio in single image super-resolution (SISR) tasks by utilizing deeper layers. However, their application is quite limited since they require high computing power. In addition, most of the existing methods rarely take full advantage of the intermediate features which are helpful for restoration. To address these issues, we propose a moderate-size SISR network named matrixed channel attention network (MCAN) by constructing a matrix ensemble of multi-connected channel attention blocks (MCAB). Several models of different sizes are released to meet various practical requirements. Conclusions can be drawn from our extensive benchmark experiments that the proposed models achieve better performance with much fewer multiply-adds and parameters. Our models will be made publicly available at this URL \footnote{\url{https://github.com/macn3388/MCAN}}.
\end{abstract}

\begin{figure*}[t]
	\newlength\fsdttwofig
	\setlength{\fsdttwofig}{-0mm}
	\scriptsize
	\centering
	\begin{tabular}{cc}
		\hspace{-0mm}
		\begin{adjustbox}{valign=t}
			\tiny
			\begin{tabular}{cccccc}
				\includegraphics[width=0.155\textwidth]{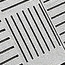} \hspace{\fsdttwofig} &
				\hspace{-.18in}
				\includegraphics[width=0.155\textwidth]{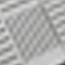} \hspace{\fsdttwofig} &
				\hspace{-.18in}
				\includegraphics[width=0.155\textwidth]{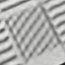} \hspace{\fsdttwofig} &
				\hspace{-.18in}
				\includegraphics[width=0.155\textwidth]{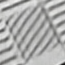} \hspace{\fsdttwofig} &
				\hspace{-.18in}
				\includegraphics[width=0.155\textwidth]{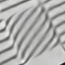} \hspace{\fsdttwofig} &
				\hspace{-.18in}
				\includegraphics[width=0.155\textwidth]{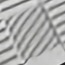} \hspace{\fsdttwofig}

				\\
								HR \hspace{\fsdttwofig} &
								\hspace{-.18in}
								Bicubic \hspace{\fsdttwofig} &
								\hspace{-.18in}
								SRCNN \cite{dong2014learning} \hspace{\fsdttwofig} &
								\hspace{-.18in}
								FSRCNN\cite{dong2016accelerating} \hspace{\fsdttwofig} &
								\hspace{-.18in}
								DRCN\cite{kim2016deeply} \hspace{\fsdttwofig} &
								\hspace{-.18in}
								VDSR\cite{kim2016accurate} 
				\\

				\includegraphics[width=0.155\textwidth]{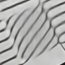} \hspace{\fsdttwofig} &
				\hspace{-.18in}
				\includegraphics[width=0.155\textwidth]{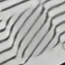} \hspace{\fsdttwofig} &
				\hspace{-.18in}
				\includegraphics[width=0.155\textwidth]{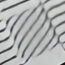} \hspace{\fsdttwofig} &
				\hspace{-.18in}
				\includegraphics[width=0.155\textwidth]{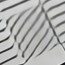} \hspace{\fsdttwofig} &
				\hspace{-.18in}
				\includegraphics[width=0.155\textwidth]{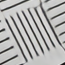} \hspace{\fsdttwofig} &
				\hspace{-.18in}
				\includegraphics[width=0.155\textwidth]{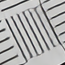} \hspace{\fsdttwofig}

				\\
												
								DRRN\cite{tai2017image} \hspace{\fsdttwofig} &
								\hspace{-.18in}
								LapSRN \cite{lai2017deep} \hspace{\fsdttwofig} &
								\hspace{-.18in}
								CARN-M \cite{ahn2018fast} \hspace{\fsdttwofig} &
								\hspace{-.18in}
								CARN \cite{ahn2018fast}  \hspace{\fsdttwofig} &
								\hspace{-.18in}
								MCAN-FAST (ours)   \hspace{\fsdttwofig} &
								\hspace{-.18in}
								MCAN (ours)
				\\
				
			\end{tabular}
		\end{adjustbox}
				\vspace{-2mm}
	\end{tabular}
	
	\caption{
		Visual results with bicubic degradation ($\times$4) on ``img\_092" from Urban100.
	}
	\label{fig:result_8x}
\end{figure*}

\section{Introduction}

Single image super-resolution (SISR) attempts to reconstruct a high-resolution (HR) image from its low-resolution (LR) equivalent, which is essentially an ill-posed inverse problem since there are infinitely many HR images that can be downsampled to the same LR image.

Most of the works discussing SISR based on deep learning have been devoted to achieving higher \emph{peak signal noise ratios} (PSNR) with deeper and deeper layers, making it difficult to fit in mobile devices \cite{kim2016accurate,kim2016deeply,tai2017image,zhang2018image}. Out of many proposals, an architecture CARN has been released that is applicable in the mobile scenario, but it is at the cost of reduction on PSNR \cite{ahn2018fast}. An information distillation network (IDN) proposed in \cite{hui2018fast} also achieves good performance at a moderate size. An effort that tackles SISR with neural architecture search has also been proposed \cite{chu2019fast,chu2019multi}, their network FALSR surpasses CARN at the same level of FLOPS.


Still, there is a noticeable gap between subjective perception and PSNR, for which a new measure called \emph{perceptual index} (PI) has been formulated \cite{blau20182018}. Noteworthy works engaging perceptual performance are SRGAN \cite{ledig2017photo} and ESRGAN \cite{wang2018esrgan}, which behave poorly on PSNR but both render more high-frequency details. However, these GAN-based methods inevitably bring about bad cases that are intolerable in practice. Our work still focuses on improving PSNR, which is a well-established distortion measure. Furthermore, our proposed model can also serve as the generator of GAN-based methods.

To seek a better trade-off between performance and applicability, we design an architecture called Matrixed Channel Attention Network. We name its basic building block \emph{multi-connected channel attention block} (MCAB), which is an adaptation of \emph{residual channel attention block} (RCAB) from RCAN \cite{zhang2018image}. MCAB differs from RCAB by allowing hierarchical connections after each activation, in such way, multiple levels of information can be passed both in depth and in breadth. 

In summary, our main contributions are as follows:
\begin{itemize}
	\item We propose a matrixed channel attention network named MCAN for SISR. It scores higher PSNR and achieves the state-of-the-art results within a lightweight range.
	\item We introduce a multi-connected channel attention block to construct matrixed channel attention cell (MCAC), which makes full use of the hierarchical features. Then we use MCAC to build a matrix-in-matrix (MIM) structure that serves as a nonlinear mapping module.
	\item We devise an edge feature fusion (EFF) block, which can be used in combination with the proposed MIM structure. The EFF can better profit the hierarchical features of MIM from the LR space.
	\item We build three additional efficient SR models of different sizes, i.e., MCAN-M, MCAN-S, MCAN-T, which are respectively short for mobile, small, and tiny. Experiments prove that all three models outperform the state-of-the-art models of the same or bigger sizes.
	\item We finally present MCAN-FAST to overcome the inefficiency of the sigmoid function on some mobile devices. Experiments show that MCAN-FAST has only a small loss of precision compared to MCAN.
\end{itemize}

\section{Related Works}
In recent years, deep learning has been applied to many areas of computer vision \cite{girshick2015fast,he2017mask,liu2016ssd,noh2015learning,zhang2016colorful}. A pioneering work  \cite{dong2014learning} has brought super-resolution into deep learning era, in which they proposed a simple three-layer convolutional neural network called SRCNN, where each layer sequentially deals with feature extraction, non-linear mapping, and reconstruction. The input of SRCNN, however, needs an extra bicubic interpolation which reduces high-frequency information and adds extra computation. Their later work FSRCNN \cite{dong2016accelerating} requires no interpolation and inserts a deconvolution layer for reconstruction, which learns an end-to-end mapping. Besides, shrinking and expanding layers are introduced to speed up computation, altogether rendering FSRCNN real-time on a generic CPU.

Meantime, VDSR presented by \cite{kim2016deeply} features a global residual learning to ease training for their very deep network. DRCN handles deep network recursively to share parameters \cite{kim2016deeply}. DRRN builds two residual blocks in a recursive manner \cite{tai2017image}. They all bear the aforementioned problem caused by interpolation. Furthermore, these very deep architectures undoubtedly require heavy computation. 

The application of DenseNet in SR domain goes to SRDenseNet \cite{tong2017image}, in which they argue that dense skip connections mitigate the vanishing gradient problem and can boost feature propagation. It achieves better performance as well as faster speed. However, results from \cite{chu2019fast} showed that dense connection might not be the most efficient and their less dense network FALSR is also competitive.

Later, a cascading residual network CARN is devised for a lightweight scenario \cite{ahn2018fast}. The basic block of their architecture is called \emph{cascading residual block}, whose outputs of intermediary layers are dispatched to each of the consequent convolutional layers. These cascading blocks, when stacked, are again organized in the same fashion. 

There is another remarkable work RCAN \cite{zhang2018image}, which has a great impact on our work. They have observed that low-frequency information is hard to capture by convolutional layers which only exploit a local region. By adding multiple long and short skip connections for residual dense blocks, low-frequency features can bypass the network and thus the main architecture focuses on high-frequency information. They also invented a \emph{channel attention} mechanism via global average pooling to deal with interdependencies among channels.

\section{Matrixed Channel Attention Network}

\subsection{Network Structure}
\begin{figure*}[ht]
	\centering
	\includegraphics[scale=0.56]{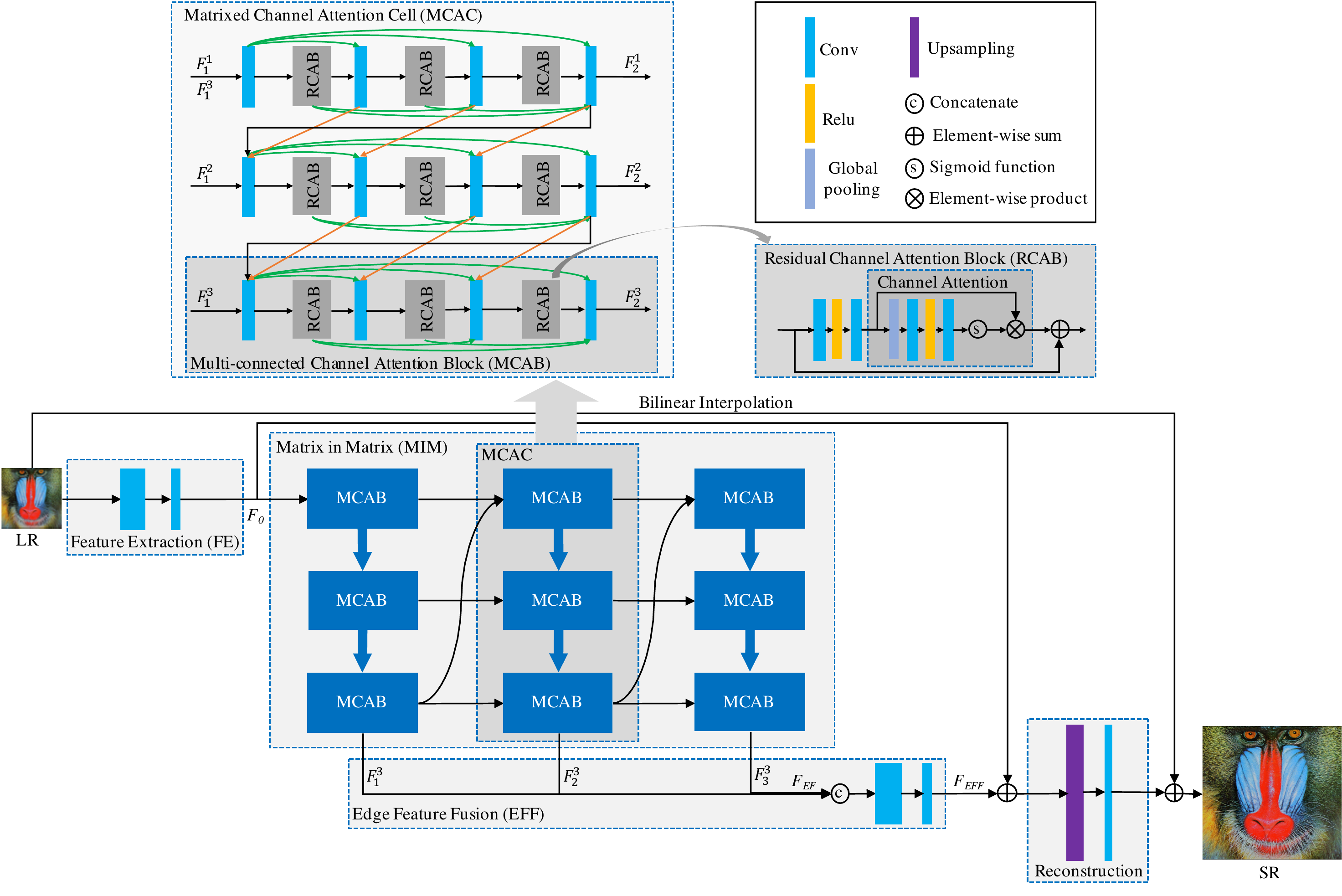}
	\vspace{-2mm}
	\caption{The architecture of matrixed convolutional neural network (MCAN), in which MIM is set to $D=3, K=3, M=3$. (The blue thick arrows indicate multiple connections between two blocks).}
	\label{fig:complete-matrixed}
\end{figure*}

The MCAN consists of four components: feature extraction (FE), matrix in matrix (MIM) mapping, edge feature fusion (EFF) and reconstruction, as illustrated in Figure~\ref{fig:complete-matrixed}.

Specifically, we utilize two successive $3 \times 3$ convolutions to extract features from the input image in the FE stage. Let $I_{LR}$ represent the input image and $I_{HR}$ be the output, this procedure can then be formulated as 
\begin{align}
\begin{split}
F_{0} &= H_{FE}(I_{LR})
\end{split}
\end{align}
where $ H_{FE}(\cdot)$ is the feature extraction function and $F_{0}$ denotes the output features. 

The nonlinear mapping is constructed by what we call a matrix-in-matrix module (MIM). Similarly,
\begin{align}
\begin{split}
F_{EF} &= H_{MIM}(F_{0})
\end{split}
\end{align}
where $H_{MIM}(\cdot)$ is the mapping function, to be discussed in detail in Section ~\ref{sec:MIM}. $F_{EF}$ stands for the edge features, so named for they are coming from the edge of our matrix block. Further feature fusion can be put formally as,
\begin{align}
\begin{split}
F_{EFF} &= H_{EFF}(F_{EF}),
\end{split}
\end{align}
We will elaborate on EFF in Section~\ref{sec:EFF}. 

Lastly, we upscale the combination of fused feature $F_{EFF}$ and $F_{0}$to generate the high-resolution target via reconstruction,
\begin{align}
\begin{split}
I_{SR} &= H_{UP}(F_{EFF} + F_{0}) + U(I_{LR}),
\end{split}
\end{align}
where $H_{UP}(\cdot)$ denotes the upsampling function and $U(\cdot)$ the bilinear interpolation.

\subsection{Matrix in Matrix}\label{sec:MIM}
The MIM block contains $D=3$ matrixed channel attention cells (MCAC). A single MCAC is 
again a sequence of $M=3$ MCABs. The $d$-th MCAC relays intermediate features to the next MCAC. In fact, each MCAC contains $K=3$ heads, which are fed into different parts of the next MCAC. We recursively define $F_d$ as the outputs of a MCAC,
\begin{align}
\begin{split}
F_{d} & = H_{MCAC}^d(F_{d-1}) \\
&= H_{MCAC}^d((F_{d-1}^1, F_{d-1}^2,\dots, F_{d-1}^K)) \\
& = (F_{d}^1, F_{d}^2,\dots, F_{d}^K). \\
\end{split}
\end{align}
Thence the output of $H_{MIM}(\cdot)$ can be composed by the combination of $K$-th outputs of all MCACs,
\begin{align}
\begin{split}
F_{EF} &= (F_{1}^K, F_{2}^K,\dots, F_{D}^K). \\
\end{split}
\end{align}

Therefore, we can regard MIM as a $D \times K$ matrix. If we look at its structural detail, a MCAC can again be seen as a matrix of $K \times M$, for this reason we call it matrix-in-matrix. The overall structure of MIM is shown in Figure \ref{fig:complete-matrixed}.

\subsection{Matrixed Channel Attention Cell}\label{sec:MCAC}
In super-resolution, skip connections are popular since it reuses intermediate features while relieving the training for deep networks \cite{kim2016deeply,mao2016image,tong2017image}. 
Nevertheless, these skip connections between modules are point-to-point, where only the output features of a module can be reused, losing many intermediate features. This can be alleviated by adding skip connections within the module, but as more intermediate features are concatenated, channels become very thick before fusion \cite{zhang2018residual}, which narrows transmission of information and gradients. 

If we densely connect all intermediate features and modules like the SRDenseNet \cite{tong2017image}, it inevitably brings in redundant connections for less important features, while the important ones become indistinguishable and increases the training difficulty.

To address these problems, we propose a matrixed channel attention cell, which is composed of several multi-connected channel attention blocks. 

\subsubsection{Multi-connected Channel Attention Block}

Previous works seldom discriminate feature channels and treat them equally. Till recently a channel attention mechanism using global pooling is proposed in RCAN to concentrate on more useful channels \cite{zhang2018image}. We adopt the same channel attention block RCAB as in RCAN, also depicted in Figure \ref{fig:complete-matrixed}, and the difference of the two only lies in the style of connections.

\textbf{Channel Attention Mechanism.} We let $X=[x_{1},\cdots,x_{c},\cdots,x_{C}]$ denote an input that contains $C$ feature maps, and the shape of each feature map be $H \times W$. Then the statistic $z_{c}$ of $c$-th feature map $x_{c}$ is defined as
	\begin{equation}
	\begin{split}
	z_{c} &= H_{GP}(x_{c}) = \dfrac{\sum_{i=1}^{H}\sum_{j=1}^{W} x_{c}(i,j)}{H \times W} , \\
	\end{split}
	\end{equation}
where $x_{c}(i,j)$ denotes the value at index $(i,j)$ of feature map $x_{c}$, and $H_{GP}(\cdot)$ represents the global average pooling function. The channel attention of the feature map $x_c$ can thus be denoted as
\begin{align}
\begin{split}
s_c &= f(W_{U}\delta(W_{D}z_c)),
\end{split}
\end{align}
where $f(\cdot)$ and $\delta(\cdot)$ represent the sigmoid function and the ReLU \cite{nair2010rectified} function respectively, $W_{D}$ is the weight set of a $1 \times 1$ convolution for channel downscaling. This convolution reduces the number of channels by a factor $r$. Later after being activated by a ReLU function, it enters a $1\times 1$ convolution for channel upscaling with the weights $W_{U}$, which expands the channel again by the factor $r$. The computed statistic $s_{c}$ is used to rescale the input features ${x_{c}}$,
\begin{align}
\begin{split}
\hat{x}_{c} &= s_{c}\cdot x_{c},
\end{split}
\end{align}

\textbf{Description of RCAB.} The RCAB is organized using the aforementioned channel attention mechanism. Formally it can be seen as a function $H_{RCAB}(\cdot)$ on the input features $I$,
\begin{align}
\begin{split}
F_{RCAB} &= H_{RCAB}(I)\\
&= s_{X_{I}}\cdot X_{I} + I \\
&= \hat{X}_{I} + I,
\end{split}
\end{align}
where $s_{X_{I}}$ is the output of channel attention on $X_{I} $, which are the features generated from the two stacked convolution layers,
\begin{align}
\begin{split}
X_{I} &= W_{2}\delta (W_{1}I).
\end{split}
\end{align}


The cascading mechanism from CARN \cite{ahn2018fast} makes use of intermediate features in a dense way. In order to relax the redundancy of dense skip connections, our residual channel attention blocks are built in a multi-connected fashion, so called as MCAB, as shown in Figure \ref{fig:complete-matrixed}. Each MCAB contains $M$ residual channel attention blocks (RCAB) and $M+1$ pointwise convolution operations for feature fusion ($H_{F}$), which are interleaved one after another.



\subsubsection{MCAC Structure}


The structure of MCAC can be considered as a matrix of $K=3$ MCABs times $M=3$ RCABs. In the $d$-th MCAC, we let the input and output of $k$-th MCAB be ${IM}_d^k$ and ${OM}_d^k$, 
and the $k$-th output of the last MCAC be $F_{d-1}^k$, we formulate ${IM}_d^k$ as follows,
\begin{align}
&\left\{
\begin{array}{lr}
\text{$[F_{0}]$} & d=k=0 \\
\text{$[{OM}_d^{k-1}]$} & d=0, k \in (0,K]\\
\text{$[F_{d-1}^k]$} &  d \in (0,D], k=0 \\
\text{$[{OM}_d^{k-1}, F_{d-1}^k]$} & d \in (0,D], k \in (0,K] \\
\end{array}
\right.
\end{align}

In the case of $d \in (0,D], k \in (0,K], m \in (0,M]$, the $m$-th feature fusion convolution $H_{F}$ takes multiple inputs and fuses them into $F_d^{k,m}$. Let the input of $m$-th RCAB be ${IR}_d^{k,m}$ and the output ${OR}_d^{k,m}$, we can write the input of $m$-th feature fusion convolution ${IF}_d^{k,m}$ as,
\begin{align}
\left\{
\begin{array}{lr}
\text{$[$}F_{d-1}^k, F_d^{k-1,m+1}, F_d^{k-1,M+1}\text{$]$} &m = 0 \\ 
\text{$[$}{OR}_d^{k,m-1}, F_d^{k-1,m+1},F_d^{k,1}, \\
\hspace{0.5cm}\dots,F_d^{k,m-1}\text{$]$}  &m \in (0,M]\\
\text{$[$}{OR}_d^{k,m-1}, F_d^{k,1},\dots, F_d^{k,M}\text{$]$} & m = M + 1\\
\end{array}
\right.
\end{align}
Now we give the complete definition of the output of $d$-th MCAC,
\begin{align}
\begin{split}
F_{d} &= (F_{d}^1, F_{d}^2,\dots, F_{d}^K)\\
&= H_{MCAC,d}(F_{d-1}) \\
&= H_{MCAC,d}((F_{d-1}^1, F_{d-1}^2,\dots, F_{d-1}^K)) \\
&= (F_d^{1,M+1},F_d^{2,M+1},\dots,F_d^{K,M+1}).
\end{split}
\end{align}

As mentioned above, the nonlinear mapping module of our proposed model can be seen as a matrix of $D \times K$. 
Thus its overall number of sigmoid functions can be calculated as,
 \begin{align}
 \begin{split}
 num_{f(\cdot)} &= D\times K \times M \times N_{ca} , 
 \end{split}
 \end{align}
where $f(\cdot)$ means the sigmoid function and $N_{ca}$ indicates the number of filters in the channel attention mechanism.

 \begin{table*}[t]
 	\vskip 0.15in
 	\scriptsize
 	\begin{center}
 		\begin{tabular}{| l| c | c | c | c | c | c | c | c |}
 			\hline
 			\multirow{2}{*}{Method} & \multirow{2}{*}{Scale} & \multirow{2}{*}{Train data} & \multirow{2}{*}{Mult-Adds} & \multirow{2}{*}{Params} & Set5 & Set14 & B100 & Urban100\\
 			\cline{6-9}
 			& & & & & PSNR/SSIM & PSNR/SSIM & PSNR/SSIM & PSNR/SSIM \\
 			\hline
 			SRCNN \cite{dong2014learning} & $\times$2 & G100+Yang91 & 52.7G & 57K & 36.66/0.9542 & 32.42/0.9063 & 31.36/0.8879 & 29.50/0.8946 \\
 			FSRCNN \cite{dong2016accelerating} & $\times$2 & G100+Yang91 & 6.0G & 12K & 37.00/0.9558 & 32.63/0.9088 & 31.53/0.8920 & 29.88/0.9020 \\
 			VDSR \cite{kim2016accurate} & $\times$2 & G100+Yang91 & 612.6G & 665K & 37.53/0.9587 & 33.03/0.9124 & 31.90/0.8960 & 30.76/0.9140 \\
 			DRCN \cite{kim2016deeply} & $\times$2 & Yang91 & 17,974.3G & 1,774K & 37.63/0.9588& 33.04/0.9118& 31.85/0.8942& 30.75/0.9133 \\
 			LapSRN \cite{lai2017deep} & $\times$2 & G200+Yang91 & 29.9G & 813K & 37.52/0.9590& 33.08/0.9130& 31.80/0.8950 & 30.41/0.9100 \\
 			DRRN \cite{tai2017image} & $\times$2 & G200+Yang91 & 6,796.9G & 297K & 37.74/0.9591 & 33.23/0.9136 & 32.05/0.8973 & 31.23/0.9188 \\
 			BTSRN \cite{fan2017balanced} & $\times$2 & DIV2K & 207.7G & 410K & 37.75/-\HSE & 33.20/-\HSE & 32.05/-\HSE & 31.63/-\HSE \\
 			MemNet \cite{tai2017memnet} & $\times$2 & G200+Yang91 & 2,662.4G & 677K & 37.78/0.9597 & 33.28/0.9142 & 32.08/0.8978 & 31.31/0.9195 \\
 			SelNet \cite{choi2017deep} & $\times$2 & ImageNet subset & 225.7G & 974K & 37.89/0.9598 & 33.61/0.9160 & 32.08/0.8984 & -\\
 			CARN \cite{ahn2018fast} & $\times$2 & DIV2K & 222.8G & 1,592K & 37.76/0.9590 & 33.52/0.9166& 32.09/0.8978 & 31.92/0.9256\\
 			CARN-M \cite{ahn2018fast} & $\times$2 & DIV2K & 91.2G & 412K & 37.53/0.9583 & 33.26/0.9141 & 31.92/0.8960 & 31.23/0.9194\\
 			MoreMNAS-A \cite{chu2019multi} & $\times$2 & DIV2K & 238.6G & 1,039K & 37.63/0.9584 & 33.23/0.9138 & 31.95/0.8961 & 31.24/0.9187\\
 			FALSR-A \cite{chu2019fast} & $\times$2 & DIV2K & 234.7G & 1,021K & 37.82/0.9595 & 33.55/0.9168 & 32.12/0.8987 & 31.93/0.9256\\
 			MCAN (ours) & $\times$2 & DIV2K & 191.3G & 1,233K & 37.91/0.9597 & 33.69/0.9183 & 32.18/0.8994 & 32.46/0.9303 \\
 			MCAN+ (ours) & $\times$2 & DIV2K & 191.3G & 1,233K & \textcolor{red}{38.10}/\textcolor{red}{0.9601} & \textcolor{red}{33.83}/\textcolor{red}{0.9197} & \textcolor{red}{32.27}/\textcolor{red}{0.9001} & \textcolor{red}{32.68}/\textcolor{red}{0.9319} \\
 			MCAN-FAST (ours) & $\times$2 & DIV2K & 191.3G & 1,233K & 37.84/0.9594 & 33.67/0.9188 & 32.16/0.8993 & 32.36/0.9300 \\
 			MCAN-FAST+ (ours) & $\times$2 & DIV2K & 191.3G & 1,233K & \textcolor{blue}{38.05}/\textcolor{blue}{0.9600} & \textcolor{blue}{33.78}/\textcolor{blue}{0.9196} & \textcolor{blue}{32.26}/\textcolor{blue}{0.8999} & \textcolor{blue}{32.62}/\textcolor{blue}{0.9317} \\
 			MCAN-M (ours) & $\times$2 & DIV2K & 105.50G & 594K & 37.78/0.9592 & 33.53/0.9174 & 32.10/0.8984 & 32.14/0.9271 \\
 			MCAN-M+ (ours) & $\times$2 & DIV2K & 105.50G & 594K & 37.98/0.9597 & 33.68/0.9186 & 32.200.8992 & 32.35/0.9290 \\
 			MCAN-S (ours) & $\times$2 & DIV2K & 46.09G & 243K & 37.62/0.9586 & 33.35/0.9156 & 32.02/0.8976 & 31.83/0.9244 \\
 			MCAN-S+ (ours) & $\times$2 & DIV2K & 46.09G & 243K & 37.82/0.9592 & 33.49/0.9168 & 32.12/0.8983 & 32.03/0.9262 \\
 			MCAN-T (ours) & $\times$2 & DIV2K & 6.27G & 35K & 37.24/0.9571 & 32.97/0.9112 & 31.74/0.8939 & 30.62/0.9120 \\
 			MCAN-T+ (ours) & $\times$2 & DIV2K & 6.27G & 35K & 37.45/0.9578 & 33.07/0.9121 & 31.85/0.8950 & 30.79/0.9137 \\
 			\hline
 			\hline
 			SRCNN \cite{dong2014learning} & $\times$3 & G100+Yang91 &52.7G & 57K &32.75/0.9090 & 29.28/0.8209 & 28.41/0.7863 & 26.24/0.7989 \\
 			FSRCNN \cite{dong2016accelerating} & $\times$3 & G100+Yang91& 5.0G & 12K & 33.16/0.9140 & 29.43/0.8242 & 28.53/0.7910 & 26.43/0.8080 \\
 			VDSR \cite{kim2016accurate} & $\times$3 & G100+Yang91 &612.6G & 665K & 33.66/0.9213 & 29.77/0.8314 & 28.82/0.7976 & 27.14/0.8279 \\
 			DRCN \cite{kim2016deeply} & $\times$3 & Yang91& 17,974.3G & 1,774K & 33.82/0.9226 & 29.76/0.8311 & 28.80/0.7963 & 27.15/0.8276 \\
 			DRRN \cite{tai2017image} & $\times$3 & G200+Yang91 & 6,796.9G & 297K & 34.03/0.9244 & 29.96/0.8349 & 28.95/0.8004 & 27.53/0.8378 \\
 			BTSRN \cite{fan2017balanced} & $\times$3 & DIV2K & 207.7G & 410K & 37.75/-\HSE & 33.20/-\HSE & 32.05/-\HSE & 31.63/-\HSE    \\
 			MemNet \cite{tai2017memnet} & $\times$3 & G200+Yang91 & 2,662.4G & 677K & 34.09/0.9248 & 30.00/0.8350 & 28.96/0.8001 & 27.56/0.8376 \\
 			SelNet \cite{choi2017deep} & $\times$3 & ImageNet subset & 120.0G & 1,159K & 34.27/0.9257 & 30.30/0.8399 & 28.97/0.8025 & - \\
 			CARN \cite{ahn2018fast} & $\times$3 & DIV2K & 118.8G & 1,592K & 34.29/0.9255 & 30.29/0.8407 & 29.06/0.8034 & 28.06/0.8493\\
 			CARN-M \cite{ahn2018fast} & $\times$3 & DIV2K & 46.1G & 412K & 33.99/0.9236 & 30.08/0.8367 & 28.91/0.8000 & 27.55/0.8385\\
 			MCAN (ours) & $\times$3 & DIV2K & 95.4G & 1,233K & 34.45/0.9271 & 30.43/0.8433 & 29.14/0.8060 & 28.47/0.8580 \\
 			MCAN+ (ours) & $\times$3 & DIV2K & 95.4G & 1,233K & \textcolor{red}{34.62}/\textcolor{red}{0.9280} & \textcolor{red}{30.50}/\textcolor{red}{0.8442} & \textcolor{red}{29.21}/\textcolor{red}{0.8070} & \textcolor{red}{28.65}/\textcolor{red}{0.8605} \\
 			MCAN-FAST (ours) & $\times$3 & DIV2K & 95.4G & 1,233K & 34.41/0.9268 & 30.40/0.8431 & 29.12/0.8055 & 28.41/0.8568 \\
 			MCAN-FAST+ (ours) & $\times$3 & DIV2K & 95.4G & 1,233K & \textcolor{blue}{34.54}/\textcolor{blue}{0.9276} & \textcolor{blue}{30.48}/\textcolor{blue}{0.8440} & \textcolor{blue}{29.20}/\textcolor{blue}{0.8067} & \textcolor{blue}{28.60}/\textcolor{blue}{0.8595} \\
 			MCAN-M (ours) & $\times$3 & DIV2K & 50.91G & 594K & 34.35/0.9261 & 30.33/0.8417 & 29.06/0.8041 & 28.22/0.8525 \\
 			MCAN-M+ (ours) & $\times$3 & DIV2K &50.91G & 594K & 34.50/0.9271 & 30.44/0.8432 & 29.14/0.8053 & 28.39/0.8552 \\
 			MCAN-S (ours) & $\times$3 & DIV2K & 21.91G & 243K & 34.12/0.9243 & 30.22/0.8391 & 28.99/0.8021 & 27.94/0.8465 \\
 			MCAN-S+ (ours) & $\times$3 & DIV2K & 21.91G & 243K & 34.28/0.9255 & 30.31/0.8403 & 29.07/0.8034 & 28.09/0.8493 \\
 			MCAN-T (ours) & $\times$3 & DIV2K & 3.10G & 35K & 33.54/0.9191 & 29.76/0.8301 & 28.73/0.7949 & 26.97/0.8243 \\
 			MCAN-T+ (ours) & $\times$3 & DIV2K & 3.10G & 35K & 33.68/0.9207 & 29.8/0.8320 & 28.80/0.7964 & 27.10/0.8271 \\
 			\hline
 			\hline
 			SRCNN \cite{dong2014learning} & $\times$4 & G100+Yang91 & 52.7G & 57K & 30.48/0.8628 & 27.49/0.7503 & 26.90/0.7101 & 24.52/0.7221 \\
 			FSRCNN \cite{dong2016accelerating} & $\times$4 & G100+Yang91& 4.6G & 12K & 30.71/0.8657 & 27.59/0.7535 & 26.98/0.7150 & 24.62/0.7280 \\
 			VDSR \cite{kim2016accurate} & $\times$4 & G100+Yang91 & 612.6G & 665K& 31.35/0.8838 & 28.01/0.7674 & 27.29/0.7251 & 25.18/0.7524 \\
 			DRCN \cite{kim2016deeply} & $\times$4 & Yang91& 17,974.3G & 1,774K & 31.53/0.8854 & 28.02/0.7670 & 27.23/0.7233 & 25.14/0.7510 \\
 			LapSRN \cite{lai2017deep} & $\times$4 & G200+Yang91 & 149.4G & 813K & 31.54/0.8850 & 28.19/0.7720 & 27.32/0.7280 & 25.21/0.7560 \\
 			DRRN \cite{tai2017image} & $\times$4 & G200+Yang91 & 6,796.9G & 297K & 31.68/0.8888 & 28.21/0.7720 & 27.38/0.7284 &25.44/0.7638 \\
 			BTSRN \cite{fan2017balanced} & $\times$4 & DIV2K & 207.7G & 410K & 37.75/-\HSE & 33.20/-\HSE & 32.05/-\HSE & 31.63/-\HSE \\
 			MemNet \cite{tai2017memnet} & $\times$4 & G200+Yang91 & 2,662.4G & 677K & 31.74/0.8893 & 28.26/0.7723 & 27.40/0.7281 & 25.50/0.7630 \\
 			SelNet \cite{choi2017deep} & $\times$4 & ImageNet subset & 83.1G & 1,417K & 32.00/0.8931 & 28.49/0.7783 & 27.44/0.7325 & -\\
 			SRDenseNet \cite{tong2017image} & $\times$4 & ImageNet subset & 389.9G & 2,015K & 32.02/0.8934 & 28.50/0.7782 & 27.53/0.7337 & 26.05/0.7819 \\
 			CARN \cite{ahn2018fast} & $\times$4 & DIV2K & 90.9G & 1,592K & 32.13/0.8937 & 28.60/0.7806 & 27.58/0.7349 & 26.07/0.7837 \\
 			CARN-M \cite{ahn2018fast} & $\times$4 & DIV2K & 32.5G & 412K & 31.92/0.8903 & 28.42/0.7762 & 27.44/0.7304 & 25.62/0.7694\\
 			MCAN (ours) & $\times$4 & DIV2K & 83.1G & 1,233K & 32.33/0.8959 & 28.72/0.7835 & 27.63/0.7378 & 26.43/0.7953 \\
 			MCAN+ (ours) & $\times$4 & DIV2K & 83.1G & 1,233K & \textcolor{red}{32.48}/\textcolor{red}{0.8974} & \textcolor{red}{28.80}/\textcolor{red}{0.7848} & \textcolor{red}{27.69}/\textcolor{red}{0.7389} & \textcolor{red}{26.58}/\textcolor{red}{0.7981}\\
 			MCAN-FAST (ours) & $\times$4 & DIV2K & 83.1G & 1,233K & 32.30/0.8955 & 28.69/0.7829 & 27.60/0.7372 & 26.37/0.7938 \\
 			MCAN-FAST+ (ours) & $\times$4 & DIV2K & 83.1G & 1,233K & \textcolor{blue}{32.43}/\textcolor{blue}{0.8970} & \textcolor{blue}{28.78}/\textcolor{blue}{0.7843} & \textcolor{blue}{27.68}/\textcolor{blue}{0.7385} & \textcolor{blue}{26.53}/\textcolor{blue}{0.7970} \\
 			MCAN-M (ours) & $\times$4 & DIV2K & 35.53G & 594K & 32.21/0.8946 & 28.63/0.7813 & 27.57/0.7357 & 26.19/0.7877 \\
 			MCAN-M+ (ours) & $\times$4 & DIV2K &35.53G & 594K & 32.34/0.8959 & 28.72/ 0.7827 & 27.63/0.7370 & 26.34/0.7909 \\
 			MCAN-S (ours) & $\times$4 & DIV2K & 13.98G & 243K & 31.97/0.8914 & 28.48/0.7775 & 27.48/0.7324 & 25.93/0.7789 \\
 			MCAN-S+ (ours) & $\times$4 & DIV2K & 13.98G & 243K & 32.11/0.8932 & 28.57/0.7791 & 27.55/0.7338 & 26.06/0.7822 \\
 			MCAN-T (ours) & $\times$4 & DIV2K & 2.00G & 35K & 31.33/0.8812 & 28.04/0.7669 & 27.22/0.7228 & 25.12/0.7515 \\
 			MCAN-T+ (ours) & $\times$4 & DIV2K &2.00G & 35K & 31.50/0.8843 & 28.14/0.7689 & 27.29/0.7244 & 25.23/0.7548 \\
 			\hline
 		\end{tabular}
 			\vspace{-2mm}
 	\end{center}
 	\caption{Quantitative comparison with the state-of-the-art methods based on $\times$2, $\times$3, $\times$4 SR with bicubic degradation model. Red/blue text: best/second-best.} 
 	\label{tab:psnr_ssim}
 	\vspace{-4mm}
 \end{table*}
\begin{figure*}[t]
	\scriptsize
	\centering
	\begin{tabular}{cc}
		\begin{adjustbox}{valign=t}
			\tiny
			\begin{tabular}{c}
				\includegraphics[width=0.252\textwidth]{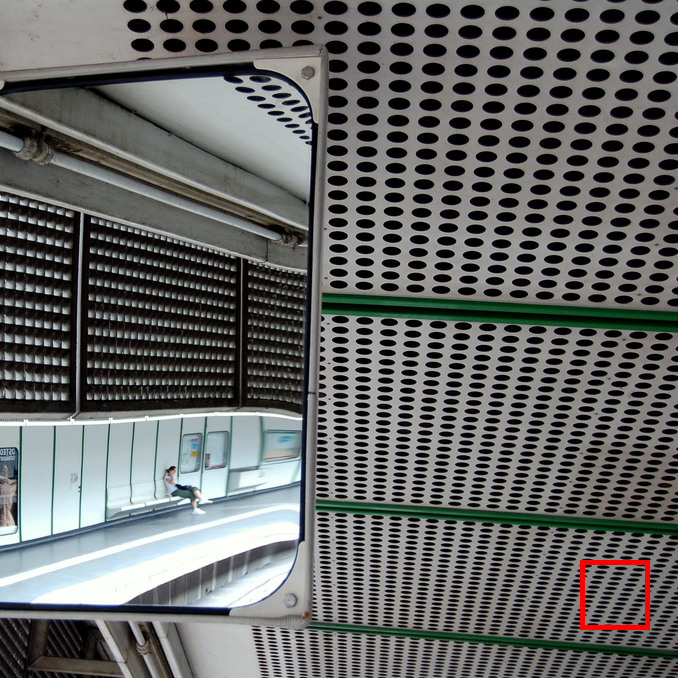}
				\\
				Urban100 ($\times$3):
				\\
				img\_004
				
			\end{tabular}
		\end{adjustbox}
		\hspace{-2.3mm}
		\begin{adjustbox}{valign=t}
			\tiny
			\begin{tabular}{cccccc}
				\hspace{-.12in}
				\includegraphics[width=0.112\textwidth]{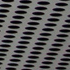} \hspace{\fsdttwofig} &
				\hspace{-.18in}
				\includegraphics[width=0.112\textwidth]{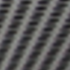} \hspace{\fsdttwofig} &
				\hspace{-.18in}
				\includegraphics[width=0.112\textwidth]{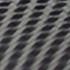} \hspace{\fsdttwofig} &
				\hspace{-.18in}
				\includegraphics[width=0.112\textwidth]{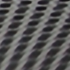} \hspace{\fsdttwofig} &
				\hspace{-.18in}
				\includegraphics[width=0.112\textwidth]{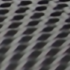} \hspace{\fsdttwofig} &
				\hspace{-.18in}
				\includegraphics[width=0.112\textwidth]{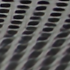} \hspace{\fsdttwofig} 
				\\				
				HR \hspace{\fsdttwofig} &
				\hspace{-.18in}
				Bicubic \hspace{\fsdttwofig} &
				\hspace{-.18in}
				FSRCNN\cite{dong2016accelerating} \hspace{\fsdttwofig} &
				\hspace{-.18in}
				VDSR\cite{kim2016accurate} &
				\hspace{-.18in}
				LapSRN \cite{lai2017deep} \hspace{\fsdttwofig} &
				\hspace{-.18in}
				CARN \cite{ahn2018fast} \hspace{\fsdttwofig}
				\\
				PSNR/SSIM \hspace{\fsdttwofig} &
				\hspace{-.18in}
				22.87/0.7859 \hspace{\fsdttwofig} &
				\hspace{-.18in}
				19.72/0.6947 \hspace{\fsdttwofig} &
				\hspace{-.18in}
				25.01/0.8841 \hspace{\fsdttwofig} &
				\hspace{-.18in}
				24.82/0.8818 \hspace{\fsdttwofig} &
				\hspace{-.18in}
				26.15/0.9086 
				\\
				\hspace{-.12in}
				\includegraphics[width=0.112\textwidth]{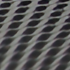} \hspace{\fsdttwofig} &
				\hspace{-.18in}
				\includegraphics[width=0.112\textwidth]{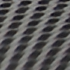} \hspace{\fsdttwofig} &
				\hspace{-.18in}
				\includegraphics[width=0.112\textwidth]{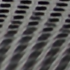} \hspace{\fsdttwofig} &
				\hspace{-.18in}
				\includegraphics[width=0.112\textwidth]{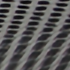} \hspace{\fsdttwofig} &
				\hspace{-.18in}
				\includegraphics[width=0.112\textwidth]{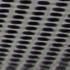} \hspace{\fsdttwofig} &
				\hspace{-.18in}
				\includegraphics[width=0.112\textwidth]{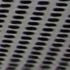} \hspace{\fsdttwofig} 
				\\
				CARN-M \cite{ahn2018fast} \hspace{\fsdttwofig} &
				\hspace{-.18in}
				MCAN-T (ours) \hspace{\fsdttwofig} &
				\hspace{-.18in}
				MCAN-S (ours) \hspace{\fsdttwofig} &
				\hspace{-.18in}
				MCAN-M (ours) \hspace{\fsdttwofig} &
				\hspace{-.18in}
				MCAN-FAST (ours) &
				\hspace{-.18in}
				MCAN (ours) \hspace{\fsdttwofig} 
				\\
				24.93/0.8898 \hspace{\fsdttwofig} &
				\hspace{-.18in}
				24.63/.8748 \hspace{\fsdttwofig} &
				\hspace{-.18in}
				26.12/0.9051 \hspace{\fsdttwofig} &
				\hspace{-.18in}
				26.51/0.9125 \hspace{\fsdttwofig} &
				\hspace{-.18in}
				\textcolor{blue}{27.83}/\textcolor{blue}{0.9259} \hspace{\fsdttwofig} &
				\hspace{-.18in}
				\textcolor{red}{27.87}/\textcolor{red}{0.9281}
				\\
				
			\end{tabular}
		\end{adjustbox}
	\end{tabular}

	\begin{tabular}{cc}
	\begin{adjustbox}{valign=t}
		\tiny
		\begin{tabular}{c}
			\includegraphics[width=0.252\textwidth]{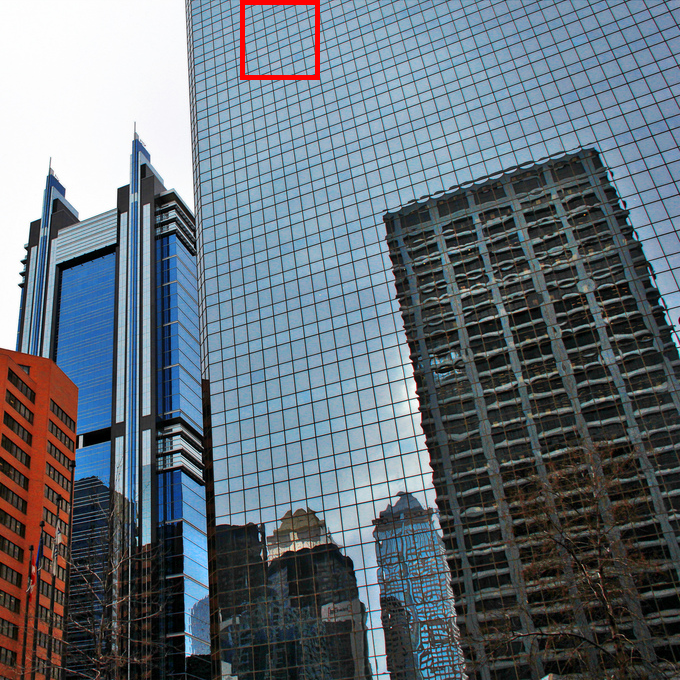}
			\\
			Urban100 ($\times$4):
			\\
			img\_099
			
		\end{tabular}
	\end{adjustbox}
	\hspace{-2.3mm}
	\begin{adjustbox}{valign=t}
		\tiny
		\begin{tabular}{cccccc}
			\hspace{-.12in}
			\includegraphics[width=0.112\textwidth]{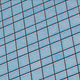} \hspace{\fsdttwofig} &
			\hspace{-.18in}
			\includegraphics[width=0.112\textwidth]{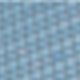} \hspace{\fsdttwofig} &
			\hspace{-.18in}
			\includegraphics[width=0.112\textwidth]{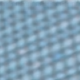} \hspace{\fsdttwofig} &
			\hspace{-.18in}
			\includegraphics[width=0.112\textwidth]{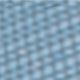} \hspace{\fsdttwofig} &
			\hspace{-.18in}
			\includegraphics[width=0.112\textwidth]{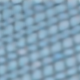} \hspace{\fsdttwofig} &
			\hspace{-.18in}
			\includegraphics[width=0.112\textwidth]{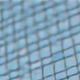} \hspace{\fsdttwofig} 
			\\				
			HR \hspace{\fsdttwofig} &
			\hspace{-.18in}
			Bicubic \hspace{\fsdttwofig} &
			\hspace{-.18in}
			FSRCNN\cite{dong2016accelerating} \hspace{\fsdttwofig} &
			\hspace{-.18in}
			VDSR\cite{kim2016accurate} &
			\hspace{-.18in}
			LapSRN \cite{lai2017deep} \hspace{\fsdttwofig} &
			\hspace{-.18in}
			CARN \cite{ahn2018fast} \hspace{\fsdttwofig}
			
			\\
			PSNR/SSIM \hspace{\fsdttwofig} &
			\hspace{-.18in}
			 22.41/0.5873 \hspace{\fsdttwofig} &
			\hspace{-.18in}
			23.61/0.6670 \hspace{\fsdttwofig} &
			\hspace{-.18in}
			24.02/0.6961 \hspace{\fsdttwofig} &
			\hspace{-.18in}
			24.47/0.7220 \hspace{\fsdttwofig} &
			\hspace{-.18in}
			25.81/0.7926
			\\
			\hspace{-.12in}
			\includegraphics[width=0.112\textwidth]{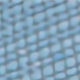} \hspace{\fsdttwofig} &
			\hspace{-.18in}
			\includegraphics[width=0.112\textwidth]{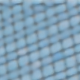} \hspace{\fsdttwofig} &
			\hspace{-.18in}
			\includegraphics[width=0.112\textwidth]{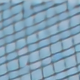} \hspace{\fsdttwofig} &
			\hspace{-.18in}
			\includegraphics[width=0.112\textwidth]{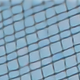} \hspace{\fsdttwofig} &
			\hspace{-.18in}
			\includegraphics[width=0.112\textwidth]{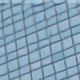} \hspace{\fsdttwofig} &
			\hspace{-.18in}
			\includegraphics[width=0.112\textwidth]{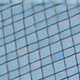} \hspace{\fsdttwofig} 
			\\
			CARN-M \cite{ahn2018fast} \hspace{\fsdttwofig} &
			\hspace{-.18in}
			MCAN-T (ours) \hspace{\fsdttwofig} &
			\hspace{-.18in}
			MCAN-S (ours) \hspace{\fsdttwofig} &
			\hspace{-.18in}
			MCAN-M (ours) \hspace{\fsdttwofig} &
			\hspace{-.18in}
			MCAN-FAST (ours) &
			\hspace{-.18in}
			MCAN (ours) \hspace{\fsdttwofig} 
			\\
			25.05/0.7557 \hspace{\fsdttwofig} &
			\hspace{-.18in}
			24.30/0.7138 \hspace{\fsdttwofig} &
			\hspace{-.18in}
			25.50/0.7793 \hspace{\fsdttwofig} &
			\hspace{-.18in}
			25.92/0.7977 \hspace{\fsdttwofig} &
			\hspace{-.18in}
			\textcolor{blue}{25.93}/\textcolor{blue}{0.8062} \hspace{\fsdttwofig} &
			\hspace{-.18in}
			\textcolor{red}{26.37}/\textcolor{red}{0.8118}
			\\
			
		\end{tabular}
	\end{adjustbox}
\end{tabular}

	\begin{tabular}{cc}
	\begin{adjustbox}{valign=t}
		\tiny
		\begin{tabular}{c}
			\includegraphics[width=0.252\textwidth]{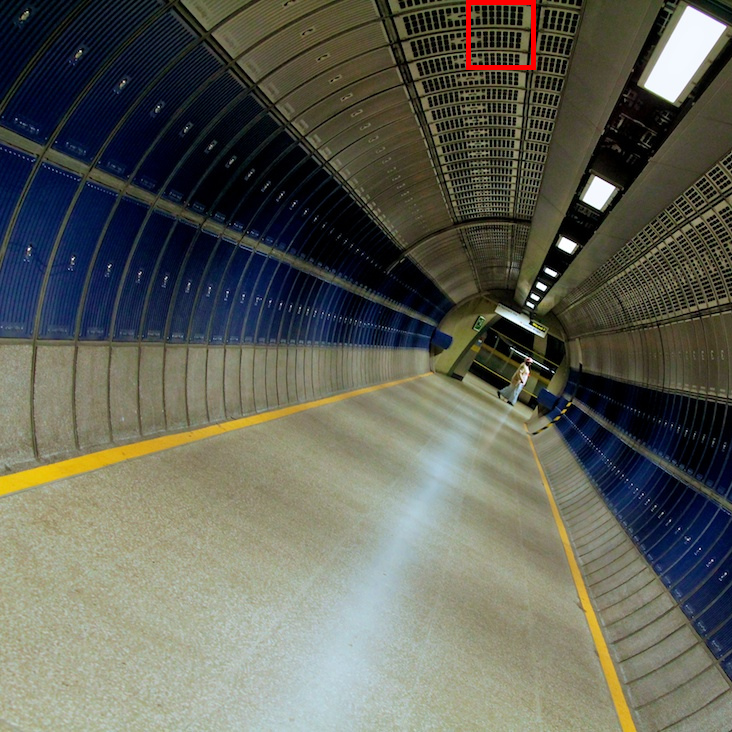}
			\\
			Urban100 ($\times$4):
			\\
			img\_078
			
		\end{tabular}
	\end{adjustbox}
	\hspace{-2.3mm}
	\begin{adjustbox}{valign=t}
		\tiny
		\begin{tabular}{cccccc}
			\hspace{-.12in}
			\includegraphics[width=0.112\textwidth]{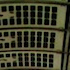} \hspace{\fsdttwofig} &
			\hspace{-.18in}
			\includegraphics[width=0.112\textwidth]{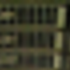} \hspace{\fsdttwofig} &
			\hspace{-.18in}
			\includegraphics[width=0.112\textwidth]{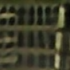} \hspace{\fsdttwofig} &
			\hspace{-.18in}
			\includegraphics[width=0.112\textwidth]{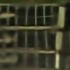} \hspace{\fsdttwofig} &
			\hspace{-.18in}
			\includegraphics[width=0.112\textwidth]{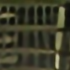} \hspace{\fsdttwofig} &
			\hspace{-.18in}
			\includegraphics[width=0.112\textwidth]{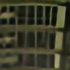} \hspace{\fsdttwofig} 
			\\				
			HR \hspace{\fsdttwofig} &
			\hspace{-.18in}
			Bicubic \hspace{\fsdttwofig} &
			\hspace{-.18in}
			FSRCNN\cite{dong2016accelerating} \hspace{\fsdttwofig} &
			\hspace{-.18in}
			VDSR\cite{kim2016accurate} &
			\hspace{-.18in}
			LapSRN \cite{lai2017deep} \hspace{\fsdttwofig} &
			\hspace{-.18in}
			CARN \cite{ahn2018fast} \hspace{\fsdttwofig}
			
			\\
			PSNR/SSIM \hspace{\fsdttwofig} &
			\hspace{-.18in}
			25.70/0.6786 \hspace{\fsdttwofig} &
			\hspace{-.18in}
			26.46/0.7338 \hspace{\fsdttwofig} &
			\hspace{-.18in}
			26.80/ 0.7545 \hspace{\fsdttwofig} &
			\hspace{-.18in}
			26.71/0.7529 \hspace{\fsdttwofig} &
			\hspace{-.18in}
			27.18/0.7716
			\\
			\hspace{-.12in}
			\includegraphics[width=0.112\textwidth]{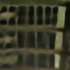} \hspace{\fsdttwofig} &
			\hspace{-.18in}
			\includegraphics[width=0.112\textwidth]{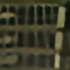} \hspace{\fsdttwofig} &
			\hspace{-.18in}
			\includegraphics[width=0.112\textwidth]{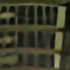} \hspace{\fsdttwofig} &
			\hspace{-.18in}
			\includegraphics[width=0.112\textwidth]{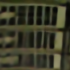} \hspace{\fsdttwofig} &
			\hspace{-.18in}
			\includegraphics[width=0.112\textwidth]{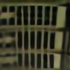} \hspace{\fsdttwofig} &
			\hspace{-.18in}
			\includegraphics[width=0.112\textwidth]{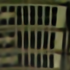} \hspace{\fsdttwofig} 
			\\
			CARN-M \cite{ahn2018fast} \hspace{\fsdttwofig} &
			\hspace{-.18in}
			MCAN-T (ours) \hspace{\fsdttwofig} &
			\hspace{-.18in}
			MCAN-S (ours) \hspace{\fsdttwofig} &
			\hspace{-.18in}
			MCAN-M (ours) \hspace{\fsdttwofig} &
			\hspace{-.18in}
			MCAN-FAST (ours) &
			\hspace{-.18in}
			MCAN (ours) \hspace{\fsdttwofig} 
			\\
			26.91/0.7597 \hspace{\fsdttwofig} &
			\hspace{-.18in}
			26.74/0.7492 \hspace{\fsdttwofig} &
			\hspace{-.18in}
			27.14/0.7678 \hspace{\fsdttwofig} &
			\hspace{-.18in}
			27.33/0.7745 \hspace{\fsdttwofig} &
			\hspace{-.18in}
			\textcolor{blue}{27.56}/\textcolor{blue}{0.7809} \hspace{\fsdttwofig} &
			\hspace{-.18in}
			\textcolor{red}{27.65}/\textcolor{red}{0.7824}
			\\
			
		\end{tabular}
	\end{adjustbox}
\end{tabular}
	\caption{
		Visual comparison with bicubic degradation model. Red/blue text: best/second-best.
	}
	\label{fig:result_8x}
\end{figure*}

\subsection{Edge Feature Fusion}\label{sec:EFF}

As we generate multiple features through MIM during different stage, we put forward an edge feature fusion (EFF) module to integrate these features hierarchically. 

Particularly, we unite the outputs of the last MCAB in each MCAC, which are nicknamed as the edge of the MIM structure. In further detail, EFF takes a $3 \times 3$ convolution for fusion and another $3\times 3$ convolution to reduce channel numbers. 
\begin{equation}
\begin{split}
F_{EFF} &=W_{F} W_{R}[F_d^{1,M+1},\dots,F_d^{K,M+1}],
\end{split}
 \end{equation}
where $W_{F}$ and $W_{R}$ are the weights of fusion convolution and the channel reduction layer.

\subsection{Comparison with Recent Models}

 \textbf{Comparison with SRDenseNet.} SRDenseNet uses dense block proposed by DenseNet to construct nonlinear mapping module \cite{tong2017image}. This dense connection mechanism may lead to redundancy, in fact not all features should be equally treated. In our work, MIM and EFF can reduce dense connections and highlight the hierarchical information. Additionally, SRDenseNet connects two blocks from point to point, which refrains transmission and utilization of intermediate features. Our proposed multi-connected channel attention block (MCAB) mitigates this problem by injecting multiple connections between blocks.
 
 \textbf{Comparison with CARN.} CARN uses a cascading block \cite{ahn2018fast}, which is also pictured in our MIM. Despite of this, MIM features multiple connections between MCACs, and the outputs of different stages are relayed between MCABs. Such an arrangement makes better use of intermediate information. Another important difference is that MCAN combines the hierarchical features before upsampling via edge feature fusion. This mechanism helps significantly for reconstruction.
 
\section{Experimental Results}
\subsection{Datasets and Evaluation Metrics}
We train our model based on DIV2K \cite{timofte2017ntire}, which contains 800 2K high-resolution images for the training set and another 100 pictures for both validation and test set. Besides, we make comparisons across three scaling tasks ($\times$2, $\times$3, $\times$4) on four datasets: Set5 \cite{bevilacqua2012low}, Set14 \cite{yang2010image}, B100 \cite{martin2001database}, and Urban100 \cite{huang2015single}. The evaluation metrics we used are PSNR \cite{hore2010image} and SSIM \cite{wang2004image} on the Y channel in the YCbCr space.

\subsection{Implementation Details}
As shown in Figure~\ref{fig:complete-matrixed}, the inputs and outputs of our model are RGB images. We crop the LR patches by $64 \times 64$ for various scale tasks and adopt the standard data augmentation. 

For training, we use Adam ($\beta_{1} = 0.9$, $\beta_{2}=0.999$, and $\epsilon=10 ^{-8}$) \cite{kingma2014adam} to minimize $L_1$ loss within $1.2 \times 10^{6}$ steps with a batch-size of $64$. The initial learning rate is set to $2\times10^{-4}$, halved every $4\times10 ^{5}$ steps. Like CARN \cite{ahn2018fast}, we also initialize the network parameters by $\theta\sim U(-k, k)$, where $k=1/\sqrt{c_{in}}$ and $c_{in}$ is the number of input feature maps. Inspired by EDSR \cite{lim2017enhanced}, we apply a multi-scale training. Our sub-pixel convolution is the same as in ESPCN \cite{shi2016real}.
\begin{table}
\centering
\begin{small}
\begin{tabular}{|c|c|c|c|c|c|} 
	\hline
	Models & $n_{FE}$     & $n_{MIM}$ & $n_{EFF}$  & $n_l$ & $r$\\
	\hline
	MCAN  & \{64,32\}  & 32 & \{96,32\} & 256 & 8 \\
	MCAN-M & \{64,24\}  & 24 & \{72,24\} & 128 & 8  \\
	MCAN-S  & \{32,16\}  & 16 & \{48,16\} & 64 & 8  \\
	MCAN-T  & \{16,8\}  & 8 & \{24,8\} & 8 & 4   \\
	\hline
\end{tabular}
\end{small}
\caption{Network hyperparameters of our networks.}
\label{tab:hyper}
\end{table}

We choose network hyperparameters to build an accurate and efficient model. The first two layers in the FE stage contain $n_{FE} = \{64, 32\}$ filters accordingly. As for MIM, we set $D=K=M=3$, its number of filters $n_{MIM} = 32$. Two EFF convolutions have $n_{EFF} = \{D \times 32, 32\}$ filters. The last convolution before the upsampling procedure has $n_{l} = 256$ filters. The reduction factor $r$ in channel attention mechanism is set to $8$. 

Since the \textit{sigmoid} function is inefficient on some mobile devices, especially for some fixed point units such as DSP. Therefore we propose MCAN-FAST by replacing the \textit{sigmoid} with the \textit{fast sigmoid} \cite{georgiou1992parallel}, which can be written as,
	\begin{equation}
	\begin{split}
	f_{fast}(x) & = \dfrac{x}{1 + \left| x \right|} . \\
	\end{split}
	\end{equation}
Experiments show that MCAN-FAST has only a small loss on precision, and it achieves almost the same level of metrics as MCAN.

For more lightweight applications, we reduce the number of filters as shown in Table \ref{tab:hyper}. Note in MCAN-T we also set the group as 4 in the group convolution of RCAB for further compression.


\subsection{Comparisons with State-of-the-art Algorithms}
\begin{figure}
	\centering
	\includegraphics[scale=0.34]{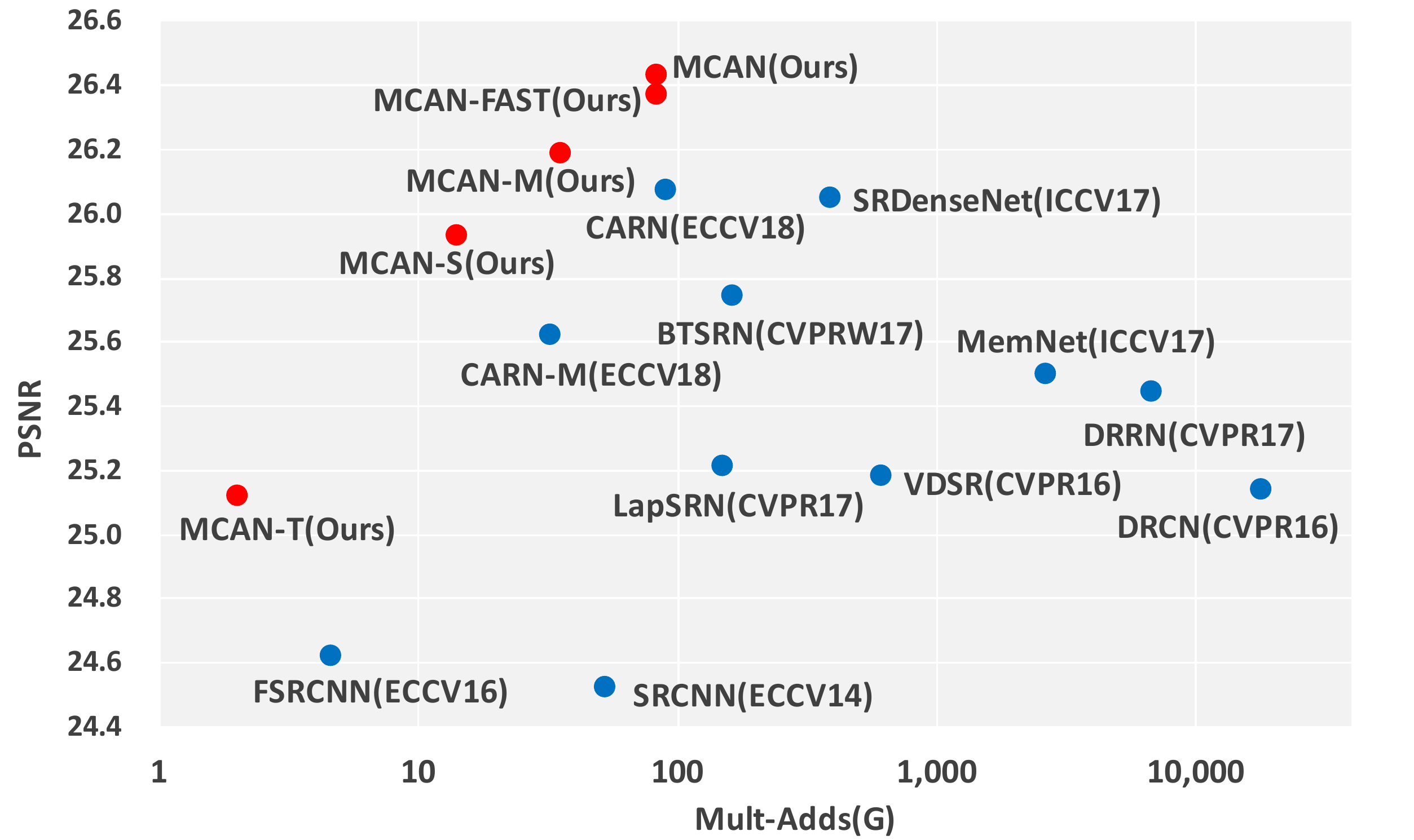}
	\caption{MCAN family (red) compared to others (blue) on $\times 4$ tasks of Urban 100. The multi-adds are calculated in the case of HR is 1280$\times$720.}
	\label{fig:scatter_plot}
\end{figure}
We use Mult-Adds and the number of parameters to measure the model size. We emphasize on Mult-Adds as it indicates the number of multiply-accumulate operations. By convention, it is normalized on $1280 \times 720$ high-resolution images. Further evaluation based on geometric self-ensembling strategy \cite{zhang2018image} are marked with `+'. 

Quantitative comparisons with the state-of-the-art methods are listed in Table~\ref{tab:psnr_ssim}. For fair comparison, we concentrate on models with comparative mult-adds and parameters.

Notably, MCAN outperforms CARN \cite{ahn2018fast} with fewer multi-adds and parameters. The medium-size model MCAN-M \cite{ahn2018fast} achieved better performance than the CARN-M, additionally, it is  still on par with CARN with about half of its multi-adds. For short, it surpasses all other listed methods, including MoreMNAS-A \cite{chu2019multi} and FALSR-A \cite{chu2019fast} from NAS methods.

The smaller model MCAN-S emulates LapSRN \cite{lai2017deep} with much fewer parameters. Particularly, it has an average advantage of $0.5$ dB on PSNR over the LapSRN on the $\times 2$ task, and on average, MCAN-S still has an advantage of $0.4$ dB. MCAN-S also behaves better than CARN-M on all tasks with half of its model size. It is worth to note that heavily compressed MCAN-S still exceeds or matches larger models such as VDSR, DRCN, DRRN and MemNet.

The tiny model MCAN-T is meant to be applied under requirements of extreme fast speed. It overtakes FSRCNN \cite{dong2016accelerating} on all tasks with the same level of mult-adds.



\subsection{Ablation Study}
\begin{table}
\centering
\begin{tabular}{|c|c|c|c|c|} 
	\hline
	MIM &\ding{56}&\ding{52}&\ding{56} &\ding{52} \\	
	\hline
	EFF &\ding{56}&\ding{56}&\ding{52}&\ding{52} \\	
	\hline
	Avg. PSNR &29.44&30.25&30.23&30.28 \\
	\hline
\end{tabular}
	\caption{Investigations of MIM and EFF. We record the best average PSNR(dB) values of Set5 $\&$ Set14 on $\times4$ SR task in $10^{5}$ steps. }
	\label{tab:ablation}
\end{table}
In this section, we demonstrate the effectiveness of the MIM structure and EFF through ablation study.

\textbf{Matrix in matrix.} We remove the connections between MCACs and also the connections between MCABs. Hence the model comes without intermediate connections. As shown in Table \ref{tab:ablation}, the MIM structure can bring significant improvements, PSNR improves from 29.44 dB to 30.25 dB when such connections are enabled. When EFF is added, PNSR continues to increase from 30.23 dB to 30.28 dB. 

\textbf{Edge feature fusion.} We simply eliminate the fusion convolutions connected to MIM and consider the output of the last MCAB as the output of MIM. In this case, the intermediate features acquired by the MIM structure are not directly involved in the reconstruction. In Table \ref{tab:ablation}, we observe that the EFF structure enhances PNSR from 29.44 dB to 30.23 dB. With MIM enabled, PSNR is further promoted from 30.25 dB to 30.28 dB. \\

%
%

\section{Conclusion}
In this paper, we proposed an accurate and efficient network with matrixed channel attention for the SISR task. Our main idea is to exploit the intermediate features hierarchically through multi-connected channel attention blocks. MCAB then acts as a basic unit that builds up the matrix-in-matrix module. We release three additional efficient models of varied sizes, MCAN-M, MCAN-S, and MCAN-T. Extensive experiments reveal that our MCAN family excel the state-of-the-art models of accordingly similar sizes or even much larger. 

To deal with the inefficiency of the sigmoid function on some mobile devices, we benefit from the fast sigmoid to construct MCAN-FAST.  The result confirms that MCAN-FAST has only a small loss of precision when compared to MCAN, and it can still achieve better performance with fewer multi-adds and parameters than the state-of-the-art methods.

{\small
\bibliographystyle{ieee}
\bibliography{iccv}
}

\end{document}